\documentclass[11pt,a4paper]{article}
\usepackage[hyperref]{acl2018}
\usepackage{times}
\usepackage{latexsym}
\usepackage{url}

\usepackage{multirow}
\usepackage{amsmath}
\usepackage{capt-of}
\usepackage{tabularx}
\usepackage[caption=false]{subfig}
\usepackage{epsfig}
\usepackage{amssymb}
\usepackage{amsfonts}
\usepackage{booktabs}
\usepackage{scalerel}
\usepackage[inline]{enumitem}
\usepackage{listings}
\usepackage{varwidth}
\usepackage[export]{adjustbox}
\usepackage{tikz}
\usetikzlibrary{tikzmark}
\usepackage{todonotes}
\usepackage{cleveref}

\usepackage{stmaryrd}
\usepackage{bbm}

\newcommand{\tabincell}[2]{\begin{tabular}{@{}#1@{}}#2\end{tabular}}

\DeclareMathOperator\softmax{\operatorname{softmax}}

\DeclareMathOperator\sigmoid{\operatorname{sigmoid}}

\DeclareMathOperator*{\argmax}{arg\,max}

\DeclareMathOperator*{\lstm}{\operatorname{f_{LSTM}}}

\usepackage{algorithm}
\usepackage[noend]{algpseudocode}

\definecolor{deepblue}{rgb}{0,0,0.5}
\definecolor{officeblue}{RGB}{0,102,204}
\definecolor{deepred}{rgb}{0.6,0,0}
\definecolor{deepgreen}{rgb}{0,0.5,0}
\definecolor{mybrickred}{RGB}{182,50,28}

\definecolor{fillcolor}{RGB}{216,217,252}

\algnewcommand\algorithmicrequireb{{\hspace{0.95cm}}}
\algnewcommand\INPTDESCB{\item[\algorithmicrequireb]}

\algnewcommand\algorithmicfuncdesc{\textbf{Function:}}
\algnewcommand\FUNCDESC{\item[\algorithmicfuncdesc]}
\algnewcommand\algorithmicfuncdescb{{\hspace{0.86cm}}}
\algnewcommand\FUNCDESCB{\item[\algorithmicfuncdescb]}
\algnewcommand{\algorithmicgoto}{\textbf{goto}}
\algnewcommand{\Goto}[1]{\algorithmicgoto~\ref{#1}}
\newcommand*\Let[2]{\State {#1 $\gets$ #2}}

\newcommand\AlgComment[1]{\color{mybrickred}{$\triangleright$ \textit{#1}}}

\newcommand\mycode[1]{\textsf{\fontsize{10}{12}\selectfont #1}}
\newcommand\base{\textsc{OneStage}}
\newcommand\ours{\textsc{Coarse2Fine}}

\lstdefinestyle{ifttt}{
	language=Python,
	otherkeywords={Trigger,Action,-,IF,THEN},             
	keywordstyle=\bfseries\color{deepblue},
	emph={MyClass,__init__},          
	emphstyle=\color{deepred},    
	showstringspaces=false,
	breaklines=true,
	escapeinside=||,
	columns=fullflexible,
	basicstyle=\fontfamily{cmtt}\small,
	belowskip=-\baselineskip,
	aboveskip=-0.7\baselineskip
}

\lstdefinestyle{django}{
	language=Python,
	otherkeywords={self},             
	keywordstyle=\bfseries\color{deepblue},
	emph={MyClass,__init__},          
	emphstyle=\color{deepred},    
	showstringspaces=false,
	breaklines=true,
	escapeinside=||,
	columns=fullflexible,
	basicstyle=\fontfamily{cmtt}\small,
	belowskip=-\baselineskip,
	aboveskip=-0.7\baselineskip
}

\lstdefinestyle{pythoncode}{
	language=Python,
	otherkeywords={self},             
	keywordstyle=\bfseries\color{deepblue},
	emph={MyClass,__init__},          
	emphstyle=\color{deepred},    
	showstringspaces=false,
	breaklines=true,
	escapeinside=||,
	columns=fullflexible,
}

\aclfinalcopy 
\def\aclpaperid{434} 

\title{Coarse-to-Fine Decoding for Neural Semantic Parsing}

\author{Li Dong \and Mirella Lapata \\
	Institute for Language, Cognition and Computation \\
	School of Informatics, University of Edinburgh \\
	10 Crichton Street, Edinburgh EH8 9AB \\
	{\tt \href{mailto:li.dong@ed.ac.uk}{li.dong@ed.ac.uk}}~~~~{\tt \href{mailto:mlap@inf.ed.ac.uk}{mlap@inf.ed.ac.uk}}}

\date{}

\begin{document}
\maketitle
\begin{abstract}
Semantic parsing aims at mapping natural language utterances into structured meaning representations. In this work, we propose a structure-aware neural architecture which decomposes the semantic parsing process into two stages. Given an input utterance, we first generate a rough sketch of its meaning, where low-level information (such as variable names and arguments) is glossed over. Then, we fill in missing details by taking into account the natural language input and the sketch itself. Experimental results on four datasets characteristic of different domains and meaning representations show that our approach consistently improves performance, achieving competitive results despite the use of relatively simple decoders.
\end{abstract}

\section{Introduction}

Semantic parsing maps natural language utterances onto machine
interpretable meaning representations (e.g., executable queries or
logical forms).  The successful application of recurrent neural
networks to a variety of NLP tasks
\cite{mt:jointly:align:translate,grammar:foreign:language} has
provided strong impetus to treat semantic parsing as a
sequence-to-sequence problem
\cite{data-recombination,lang2logic,latent-predictor}. The fact that
meaning representations are typically structured objects has prompted
efforts to develop neural architectures which explicitly account for
their structure. Examples include tree
decoders~\cite{lang2logic,tree-doubly}, decoders constrained by a
grammar model~\cite{grammar-nsp,nl2code,table-nsp}, or modular
decoders which use syntax to dynamically compose various
submodels~\cite{asn}.

In this work, we propose to decompose the decoding process into two
stages. The first decoder focuses on predicting a rough
\textit{sketch} of the meaning representation, which omits low-level
details, such as arguments and variable names. Example sketches for
various meaning representations are shown in
Table~\ref{tbl:dataset}. Then, a second decoder fills in missing
details by conditioning on the natural language input and the sketch
itself. Specifically, the sketch constrains the generation process
and is encoded into vectors to guide decoding.

We argue that there are at least three advantages to the proposed
approach. Firstly, the decomposition disentangles high-level from
low-level semantic information, which enables the decoders to model
meaning at different levels of granularity. As shown in
Table~\ref{tbl:dataset}, sketches are more compact
and as a result easier to generate compared to decoding the entire
meaning structure in one go.  Secondly,
the model can explicitly share knowledge of coarse structures
for the examples that
have the same sketch (i.e.,~basic meaning), even though their actual
meaning representations are different (e.g.,~due to different
details). Thirdly, after generating the sketch, the decoder knows what
the basic meaning of the utterance looks like, and the model can use
it as global context to improve the prediction of the final details.

\begin{table*}[t]
\centering
\small
\setlength\tabcolsep{3pt}
\begin{tabular}{l c l}
\toprule
\textbf{Dataset} & \textbf{Length} & \textbf{Example} \\ \midrule
\textsc{Geo} & \tabincell{c}{7.6 \\ 13.7 \\ 6.9} & \tabincell{l}{$x:$~\textit{which state has the most rivers running through it?} \\ {$y:$~(argmax \$0 (state:t \$0) (count \$1 (and (river:t \$1) (loc:t \$1 \$0))))} \\ {$a:$~(argmax\#1 state:t@1 (count\#1 (and river:t@1 loc:t@2 ) ) )} } \\ \midrule
\textsc{Atis} & \tabincell{c}{11.1 \\ 21.1 \\ 9.2} & \tabincell{l}{$x:$~\textit{all flights from dallas before 10am} \\{$y:$~(lambda \$0 e (and (flight \$0) (from \$0 dallas:ci) (\textless~(departure\_time \$0) 1000:ti)))} \\ {$a:$~(lambda\#2 (and flight@1 from@2 (\textless~departure\_time@1 ? ) ) )} } \\ \midrule
\textsc{Django} & \tabincell{c}{14.4 \\ 8.7 \\ 8.0} & \tabincell{l}{$x:$~\textit{if length of bits is lesser than integer 3 or second element of bits is not equal to string 'as' ,} \\ {$y:$~if len(bits)~\textless~3 or bits[1] != \textsf{'}as\textsf{'}:} \\ {$a:$~if len ( NAME )~\textless~NUMBER or NAME [ NUMBER ] != STRING :} } \\ \midrule
\textsc{WikiSQL} & \tabincell{c}{17.9 \\ 13.3 \\ 13.0 \\ 2.7} & \tabincell{l}{{Table schema: \textit{\textbardbl Pianist\textbardbl Conductor\textbardbl Record Company\textbardbl Year of Recording\textbardbl Format\textbardbl}} \\ $x:$~\textit{What record company did conductor Mikhail Snitko record for after 1996?} \\ {$y:$~\texttt{SELECT} \textit{Record Company} \texttt{WHERE} (\textit{Year of Recording}~\textgreater~\textit{1996}) \texttt{AND} (\textit{Conductor} = \textit{Mikhail Snitko})} \\ {$a:$~\texttt{WHERE}~\textgreater~\texttt{AND}~=} } \\ \bottomrule
\end{tabular}
\normalsize
\caption{Examples of natural language expressions~$x$, their meaning representations~$y$, and meaning sketches~$a$. The average number of tokens is shown in the second column.}
\label{tbl:dataset}
\end{table*}

Our framework is flexible and not restricted to specific tasks or any
particular model. We conduct experiments on four datasets
representative of various semantic parsing tasks ranging from logical
form parsing, to code generation, and SQL query generation. We adapt
our architecture to these tasks and present several ways to obtain
sketches from their respective meaning representations.  Experimental
results show that our framework achieves competitive performance
compared with previous systems, despite employing relatively simple
sequence decoders.

\section{Related Work}
\label{sec:related}

Various models have been proposed over the years to learn semantic
parsers from natural language expressions paired with their meaning
representations~\cite{tang-mooney:2000:EMNLP,scissor,zc07,lambdawasp,lnlz08,fubl,sp:as:mt,tisp}. These
systems typically learn lexicalized mapping rules and scoring models
to construct a meaning representation for a given input.

More recently, neural sequence-to-sequence models have been applied
to semantic parsing with promising
results~\cite{lang2logic,data-recombination,latent-predictor},
eschewing the need for extensive feature engineering.  Several ideas
have been explored to enhance the performance of these models such as
data augmentation~\cite{semi-nsp,data-recombination}, transfer
learning~\cite{transfer-nsp}, sharing parameters for multiple
languages or meaning
representations~\cite{multilingual-nsp,multi-kb-nsp}, and utilizing
user feedback signals~\cite{user-feedback}. There are also efforts to
develop structured decoders that make use of the syntax of meaning
representations. \citet{lang2logic} and \citet{tree-doubly} develop
models which generate tree structures in a top-down fashion.
\citet{grammar-nsp} and \citet{table-nsp} employ the grammar to constrain
the decoding process.
\citet{cheng17transition} use a transition system to generate
variable-free queries.
\citet{nl2code} design a grammar model for the
generation of abstract syntax trees~\cite{ast} in depth-first,
left-to-right order.  \citet{asn} propose a modular decoder whose
submodels are dynamically composed according to the generated tree
structure.

Our own work also aims to model the structure of meaning
representations more faithfully.
The flexibility of our approach enables us to easily apply
sketches to different types
of meaning representations, e.g.,~trees or other structured objects.
Coarse-to-fine methods have been popular in the NLP literature, and
are perhaps best known for syntactic
parsing~\cite{charniak-EtAl:2006:HLT-NAACL06-Main,petrov2011coarse}.
\citet{map:instru} and \citet{macro17yuchen} use coarse lexical entries
or macro grammars to reduce the search space of semantic parsers.
Compared with coarse-to-fine inference for
lexical induction, sketches in our case are abstractions of the final
meaning representation.

The idea of using sketches as intermediate representations has also
been explored in the field of program
synthesis~\cite{sketch08solarphd,sketch13ase,sketch17popl}.
\citet{sketch17sqlizer} use \textsc{Sempre}~\cite{webqa} to map a
sentence into SQL sketches which are completed using program synthesis
techniques and iteratively repaired if they are faulty.

\begin{figure*}[t]
\centering
\hbox{\includegraphics[width=1\textwidth]{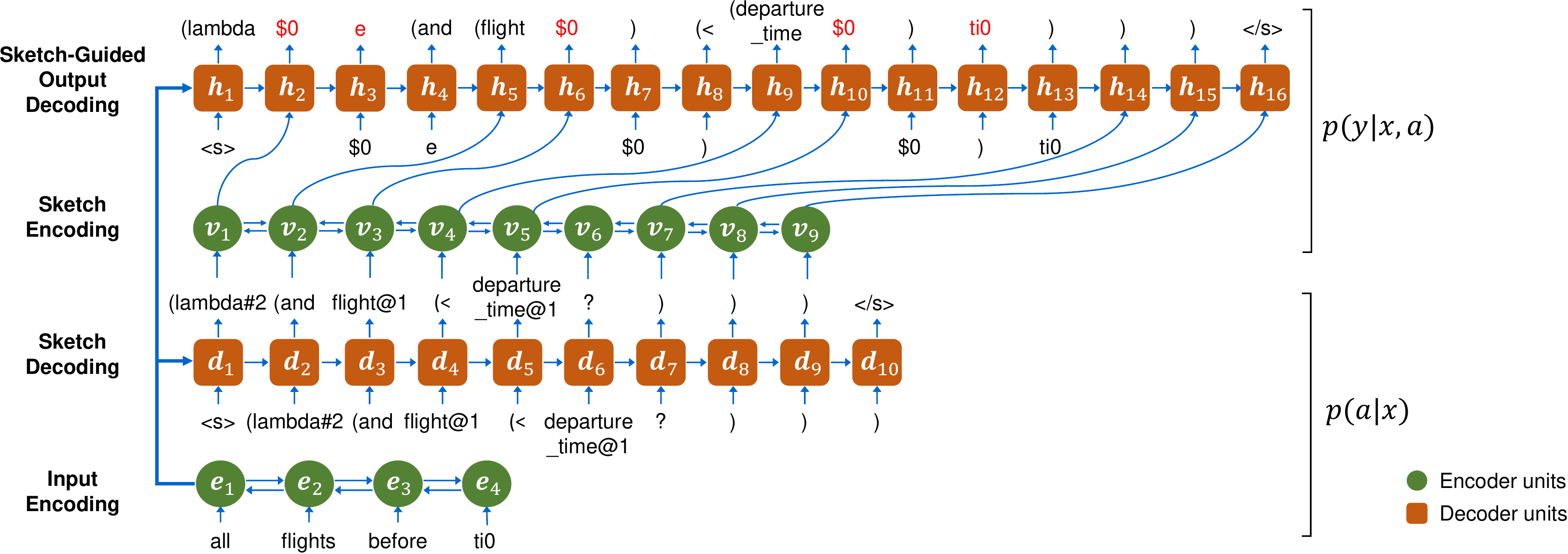}}
\caption{We first generate the meaning sketch~$a$~for natural language input~$x$.
Then, a fine meaning decoder fills in the missing details (shown in {\color{red} red}) of meaning representation~$y$. The coarse structure~$a$~is used to guide and constrain the output decoding.
}
\label{fig:method}
\end{figure*}

\section{Problem Formulation}
\label{sec:method}

Our goal is to learn semantic parsers from instances of natural
language expressions paired with their structured meaning representations.
Let~$x = x_1 \cdots x_{|x|}$ denote a natural language expression,
and~$y = y_1 \cdots y_{|y|}$ its meaning representation. We wish to
estimate~$p\left(y | x\right)$, the conditional probability of meaning
representation~$y$ given input~$x$. We decompose~$p\left(y | x\right)$
into a two-stage generation process:
\begin{equation}
\label{eq:overview}
p\left(y | x\right) = p\left(y | x, a \right) p\left(a | x\right)
\end{equation}
where $a = a_1 \cdots a_{|a|}$ is an abstract sketch representing the
meaning of~$y$. We defer detailed description of how sketches are
extracted to Section~\ref{sec:task}. Suffice it to say that the
extraction amounts to stripping off arguments and variable names in
logical forms, schema specific information in SQL queries, and
substituting tokens with types in source code (see
Table~\ref{tbl:dataset}).

As shown in Figure~\ref{fig:method}, we first predict sketch~$a$ for
input~$x$, and then fill in missing details to generate the
final meaning representation~$y$ by conditioning on both~$x$
and~$a$. The sketch is encoded into vectors which in turn guide and
constrain the decoding of~$y$. We view the input expression~$x$, the
meaning representation~$y$, and its sketch~$a$ as sequences.
The generation probabilities are factorized as:
\begin{align}
p\left( a | x \right) &= \prod_{ t = 1 }^{ |a| }{ p\left( a_t | a_{<t} , x \right) } \label{eq:prob:whole:seq:a} \\
p\left( y | x, a\right) &= \prod_{ t = 1 }^{ |y| }{ p\left( y_t | y_{<t} , x , a \right) } \label{eq:prob:whole:seq:y}
\end{align}
where $a_{<t} = a_1 \cdots a_{t-1}$, and $y_{<t} = y_1 \cdots y_{t-1}$. In the following, we will explain how $p\left(a | x\right)$ and $p\left(y | x, a \right)$ are estimated.

\subsection{Sketch Generation}
\label{sec:sketch:generation}

An \textit{encoder} is used to encode the natural language input $x$ into vector representations. Then, a \textit{decoder} learns to compute $p\left( a | x \right)$ and generate the sketch $a$ conditioned on the encoding vectors.

\paragraph{Input Encoder}
Every input word is mapped to a vector via
$\mathbf{x}_t = \mathbf{W}_x \mathbf{o}\left( x_t \right) $, where
$\mathbf{W}_x \in \mathbb{R}^{n \times |\mathcal{V}_x|}$ is an
embedding matrix, $|\mathcal{V}_x|$~is the vocabulary size, and
$\mathbf{o}\left( x_t \right)$~a one-hot vector.  We use a
bi-directional recurrent neural network with long short-term memory
units (\textsc{LSTM}, \citealt{lstm}) as the input encoder. The
encoder recursively computes the hidden vectors at the $t$-th time
step via:
\begin{align}
\overrightarrow{\mathbf{e} }_{ t } &= \lstm\left( \overrightarrow{\mathbf{e} }_{ t-1 } , \mathbf{x}_t \right) , t = 1, \cdots , |x| \label{eq:encoder:lstm:right} \\
\overleftarrow{\mathbf{e} }_{ t } &= \lstm\left( \overleftarrow{\mathbf{e} }_{ t+1 } , \mathbf{x}_t \right) , t = |x|, \cdots , 1 \label{eq:encoder:lstm:left} \\
\mathbf{e}_t &= [\overrightarrow{\mathbf{e} }_{ t }, \overleftarrow{\mathbf{e} }_{ t }] \label{eq:encoder:lstm}
\end{align}
where $[\cdot,\cdot]$ denotes vector concatenation, $\mathbf{e}_t \in \mathbb{R}^{n}$, and $\lstm$ is the LSTM function.

\paragraph{Coarse Meaning Decoder}
The decoder's hidden vector at the $t$-th time step is computed by
${\mathbf{d}}_{t} = \lstm \left( {\mathbf{d}}_{t-1},
  {\mathbf{a}}_{t-1} \right)$,
where ${\mathbf{a}}_{t-1} \in \mathbb{R}^{n}$ is the embedding of the
previously predicted token.  The hidden states of the first time step
in the decoder are initialized by the concatenated encoding vectors
${\mathbf{d}}_{0} = [\overrightarrow{\mathbf{e} }_{ |x| },
\overleftarrow{\mathbf{e} }_{ 1 }]$.
Additionally, we use an attention mechanism~\cite{luong-attention} to
learn soft alignments.  We compute the attention score for the current
time step~$t$ of the decoder, with the \mbox{$k$-th}~hidden state in
the encoder as:
\begin{equation}
\label{eq:attention:score}
{ s }_{ t,k } = { \exp \{ {\mathbf{d}}_{ t } \cdot {\mathbf{e}}_{ k } \} } / Z_t
\end{equation}
where $Z_t = \sum _{ j=1 }^{ |x| }{ \exp \{ {\mathbf{d}}_{ t } \cdot {\mathbf{e}}_{ j } \} }$ is a normalization term.
Then we compute $p\left( a_t | a_{<t} , x \right)$ via:
\begin{align}
\mathbf{e}_{t}^{d} &= \sum_{ k=1 }^{ |x| }{ { s }_{ t,k } {\mathbf{e}}_{ k } } \\
{\mathbf{d} }_{ t }^{ att } &= \tanh \left( \mathbf{W}_1 {\mathbf{d} }_{ t } + \mathbf{W}_2 \mathbf{e}_{t}^{d} \right) \label{eq:attention:new:hidden} \\
p\left( a_t | a_{<t} , x \right) &= \softmax_{a_t} \left( \mathbf{W}_o {\mathbf{d} }_{ t }^{ att } + \mathbf{b}_o \right) \label{eq:attention:decoder:predict}
\end{align}
where $\mathbf{W}_1 , \mathbf{W}_2 \in \mathbb{R}^{n \times n}$, $\mathbf{W}_o \in \mathbb{R}^{|\mathcal{V}_a| \times n}$, and $\mathbf{b}_o \in \mathbb{R}^{|\mathcal{V}_a|}$ are parameters.
Generation terminates once an end-of-sequence token
``\textit{\textless/s\textgreater}'' is emitted.

\subsection{Meaning Representation Generation}
\label{sec:mr:generation}

Meaning representations are predicted by conditioning on the input~$x$
and the generated sketch $a$. The model uses the encoder-decoder
architecture to compute $p\left( y | x, a\right)$, and decorates the
sketch $a$ with details to generate the final output.

\paragraph{Sketch Encoder}
As shown in Figure~\ref{fig:method}, a bi-directional LSTM encoder
maps the sketch sequence~$a$ into vectors
$\{\mathbf{v}_k\}_{k=1}^{|a|}$ as in Equation~\eqref{eq:encoder:lstm},
where~$\mathbf{v}_k$ denotes the vector of the $k$-th time step.

\paragraph{Fine Meaning Decoder}
The final decoder is based on recurrent neural networks with an
attention mechanism, and shares the input encoder described in
Section~\ref{sec:sketch:generation}.  The decoder's hidden states
$\{\mathbf{h}_t\}_{t=1}^{|y|}$ are computed via:
\begin{align}
\mathbf{i}_t &= \begin{cases}
\mathbf{v}_k & y_{t-1}\text{~is~determined~by~}a_k \\
{\mathbf{y}}_{t-1} & \text{otherwise}
\end{cases} \label{eq:output:decoder:i} \\
\mathbf{h}_t &= \lstm\left( \mathbf{h}_{t-1} , \mathbf{i}_t \right) \nonumber
\end{align}
where
${\mathbf{h}}_{0} = [\overrightarrow{\mathbf{e} }_{ |x| },
\overleftarrow{\mathbf{e} }_{ 1 }]$,
and ${\mathbf{y}}_{t-1}$ is the embedding of the previously predicted
token. 
Apart from using
the embeddings of previous tokens, the decoder is also fed
with~$\{\mathbf{v}_k\}_{k=1}^{|a|}$. If $y_{t-1}$ is determined
by~$a_k$ in the sketch (i.e., there is a one-to-one alignment between
$y_{t-1}$ and $a_k$), we use the corresponding token's
vector~$\mathbf{v}_k$ as input to the next time step.

The sketch constrains the decoding output. If the output token~$y_t$
is already in the sketch, we force~$y_t$ to conform to the sketch.  In
some cases, sketch tokens will indicate what information is missing
(e.g., in Figure~\ref{fig:method}, token ``\textit{flight@1}''
indicates that an argument is missing for the predicate
``\textit{flight}''). In other cases, sketch tokens will not reveal
the number of missing tokens (e.g.,~``\mycode{STRING}'' in
\textsc{Django}) but the decoder's output will indicate whether
missing details have been generated (e.g., if the decoder emits a
closing quote token for ``\mycode{STRING}'').  Moreover, type
information in sketches can be used to constrain generation. In
Table~\ref{tbl:dataset}, sketch token ``\mycode{NUMBER}'' specifies
that a numeric token should be emitted.

For the missing details, we use the hidden vector $\mathbf{h}_t$ to compute $p\left( y_t | y_{<t} , x , a \right)$, analogously
to~\Crefrange{eq:attention:score}{eq:attention:decoder:predict}.

\subsection{Training and Inference}
\label{sec:training:inference}

The model's training objective is to maximize the log likelihood of the generated meaning representations given natural language expressions:
\begin{equation}
\max \sum_{(x, a, y) \in \mathcal{D} }{ \log{p \left( y | x, a \right)} + \log{p \left( a | x \right)} } \nonumber
\end{equation}
where $\mathcal{D}$ represents training pairs.

At test time, the prediction for input $x$ is obtained via
$\hat{a} = \argmax_{a'}{ p \left( a' | x \right) }$ and
$\hat{y} = \argmax_{y'}{ p \left( y' | x, \hat{a} \right) }$, where
$a'$ and $y'$ represent coarse- and fine-grained meaning
candidates. Because probabilities $p \left( a | x \right)$ and
$p \left( y | x, a \right)$ are factorized as shown in
\Crefrange{eq:prob:whole:seq:a}{eq:prob:whole:seq:y}, we can obtain
best results approximately by using greedy search to generate tokens
one by one, rather than iterating over all candidates.

\section{Semantic Parsing Tasks}
\label{sec:task}

In order to show that our framework applies across domains and meaning
representations, we developed models for three tasks, namely parsing
natural language to logical form, to Python source code, and to SQL
query.  For each of these tasks we describe the datasets we used, how
sketches were extracted, and specify model details over and above the
architecture presented in Section~\ref{sec:method}.

\subsection{Natural Language to Logical Form}

\begin{algorithm}[t]
\caption{Sketch for \textsc{Geo} and \textsc{Atis} \label{alg:sketch}}
\begin{algorithmic}[0]
\small
\Require $t$: Tree-structure $\lambda$-calculus expression
\FUNCDESCB $t.pred$: Predicate name, or operator name
\Ensure $a$: Meaning sketch

\State{\AlgComment{(count \$0 (\textless~(fare \$0) 50:do))$\rightarrow$(count\#1 (\textless~fare@1 ?))}}
\Function{Sketch}{$t$}
\If{$t$ is leaf}\hfill{\AlgComment{No nonterminal in arguments}}
\State{\Return{``\texttt{\%}s@\texttt{\%}d'' \texttt{\%} ($t.pred , \texttt{len}\text{(}t.args$))}}
\EndIf
\If{$t.pred$ is $\lambda$ operator, or quantifier}\hfill{\AlgComment{e.g., count}}
\State{Omit variable information defined by $t.pred$}
\Let{$t.pred$}{``\texttt{\%}s\#\texttt{\%}d'' \texttt{\%} ($t.pred , \texttt{len}\text{(}variable$))}
\EndIf
\For{$c \gets$ argument in $t.args$}
\If{$c$ is nonterminal}
\Let{c}{\Call{Sketch}{c}}
\Else
\Let{c}{``?''}\hfill{\AlgComment{Placeholder for terminal}}
\EndIf
\EndFor
\State{\Return{$t$}}
\EndFunction
\normalsize
\end{algorithmic}
\end{algorithm}

For our first task we used two benchmark datasets, namely \textsc{Geo}
($880$ language queries to a database of U.S. geography) and
\textsc{Atis} ($5,410$ queries to a flight booking system). Examples
are shown in Table~\ref{tbl:dataset} (see the first and second block). We used
standard splits for both datasets: $600$ training and $280$ test
instances for \textsc{Geo}~\cite{zc05}; $4,480$ training, $480$
development, and $450$ test examples for \textsc{Atis}.  Meaning
representations in these datasets are based on
\mbox{$\lambda$-calculus}~\cite{fubl}. We use brackets to linearize
the hierarchical structure. The first element between a pair of
brackets is an operator or predicate name, and any remaining elements
are its arguments.

Algorithm~\ref{alg:sketch} shows the pseudocode used to extract
sketches from $\lambda$-calculus-based meaning representations.  We
strip off arguments and variable names in logical forms, while keeping
predicates, operators, and composition information. We use the symbol
``\textit{@}'' to denote the number of missing arguments in a
predicate. For example, we extract ``\textit{from@2}'' from the expression
``\textit{(from \$0 dallas:ci)}'' which indicates that the predicate
``\textit{from}'' has two arguments. We use ``\textit{?}'' as a
placeholder in cases where only partial argument information can be
omitted. We also omit variable information defined by the lambda
operator and quantifiers (e.g.,~\textit{exists}, \textit{count}, and
\textit{argmax}). We use the symbol ``\textit{\#}'' to denote the number
of omitted tokens. For the example in Figure~\ref{fig:method},
``\textit{lambda \$0 e}'' is reduced to ``\textit{lambda\#2}''.

The meaning representations of these two datasets are highly
compositional, which motivates us to utilize the hierarchical
structure of \mbox{$\lambda$-calculus}. A similar idea is also
explored in the tree decoders proposed in~\citet{lang2logic}
and~\citet{nl2code} where parent hidden states are fed to the input
gate of the LSTM units. On the contrary, parent hidden states serve as
input to the softmax classifiers of both fine and coarse meaning decoders.

\paragraph{Parent Feeding}
Taking the meaning sketch ``\textit{(and flight@1 from@2)}'' as an example, the parent of ``\textit{from@2}'' is ``\textit{(and}''. Let $p_t$ denote the parent of the $t$-th time step in the decoder. Compared with Equation~\eqref{eq:attention:decoder:predict}, we use the vector ${\mathbf{d} }_{ t }^{ att }$ and the hidden state of its parent ${\mathbf{d} }_{ p_t }$ to compute the probability $p\left( a_t | a_{<t} , x \right)$ via:
\begin{equation}
p\left( a_t | a_{<t} , x \right) = \softmax_{a_t} \left( \mathbf{W}_o [ {\mathbf{d} }_{ t }^{ att } , {\mathbf{d} }_{ p_t } ] + \mathbf{b}_o \right) \nonumber
\end{equation}
where $[\cdot,\cdot]$ denotes vector concatenation. The parent feeding is used for both decoding stages.

\subsection{Natural Language to Source Code}

Our second semantic parsing task used \textsc{Django} \cite{django}, a
dataset built upon the Python code of the Django library. The dataset
contains lines of code paired with natural language expressions (see
the third block in Table~\ref{tbl:dataset}) and exhibits a variety of
use cases, such as iteration, exception handling, and string
manipulation. The original split has $16,000$ training, $1,000$
development, and $1,805$ test instances.

We used the built-in lexical scanner
of Python\footnote{\url{https://docs.python.org/3/library/tokenize}}
to tokenize the code and obtain token types.  Sketches were extracted
by substituting the original tokens with their token types, except
delimiters (e.g.,~``\mycode{[}'', and ``\mycode{:}''), operators
(e.g.,~``\mycode{+}'', and ``\mycode{*}''), and built-in keywords
(e.g.,~``\mycode{True}'', and ``\mycode{while}''). For instance, the
expression ``\mycode{if s[:4].lower() == 'http':}'' becomes
``\mycode{if NAME [ : NUMBER ] . NAME ( ) == STRING :}'', with details
about names, values, and strings being omitted.

\textsc{Django} is a diverse dataset, spanning various real-world use
cases and as a result models are often faced with out-of-vocabulary
(OOV) tokens (e.g.,~variable names, and numbers) that are unseen
during training. We handle OOV tokens with a copying
mechanism~\cite{copynet,gulcehre16copy,data-recombination}, which allows the fine
meaning decoder (Section~\ref{sec:mr:generation}) to directly copy
tokens from the natural language input.

\paragraph{Copying Mechanism}
Recall that we use a softmax classifier to predict the probability
distribution $p\left( y_t | y_{<t} , x , a \right)$ over the
pre-defined vocabulary. We also learn a copying gate $g_t \in [0,1]$
to decide whether~$y_t$ should be copied from the input or generated from
the vocabulary. We compute the modified output distribution via:
\begin{align}
g_t &= \sigmoid( \mathbf{w}_{g} \cdot \mathbf{h}_{t} + b_g ) \nonumber \\
\tilde{p}\left( y_t | y_{<t} , x , a \right) &= (1 - g_t) p\left( y_t | y_{<t} , x , a \right) \nonumber \\
&~~~~~~+ \mathbbm{1}_{[y_t \notin \mathcal{V}_{y}]} g_t \hspace*{-1ex}\sum_{k: x_k = y_t}\hspace*{-1ex}{s_{t,k}} \nonumber
\end{align}
where $\mathbf{w}_{g} \in \mathbb{R}^{n}$ and $b_g \in \mathbb{R}$ are parameters,
and the indicator function $\mathbbm{1}_{[y_t \notin \mathcal{V}_{y}]}$
is $1$ only if $y_t$ is not in the target vocabulary $\mathcal{V}_{y}$;
the attention score $s_{t,k}$ (see
Equation~\eqref{eq:attention:score}) measures how likely it is to
copy~$y_t$ from the input word~$x_k$.

\subsection{Natural Language to SQL}
\label{sec:method:wikisql}

The \textsc{WikiSQL}~\cite{wikisql} dataset contains~$80,654$
examples of questions and SQL queries distributed
across~$24,241$ tables from Wikipedia. The goal is to generate the
correct SQL query for a natural language question and table schema
(i.e.,~table column names), without using the content values of tables
(see the last block in Table~\ref{tbl:dataset} for an example). The
dataset is partitioned into a training set ($70\%$), a development set
($10\%$), and a test set ($20\%$). Each table is present in one split
to ensure generalization to unseen tables.

\textsc{WikiSQL} queries follow the format ``\texttt{SELECT}
\texttt{agg\_op} \texttt{agg\_col} \texttt{WHERE} (\texttt{cond\_col}
\texttt{cond\_op} \texttt{cond}) \texttt{AND} ...'', which is a subset
of the SQL syntax.  \texttt{SELECT} identifies the column that is to
be included in the results after applying the aggregation operator
\texttt{agg\_op}\footnote{\texttt{agg\_op} $\in \{ empty,
  \texttt{COUNT}, \texttt{MIN}, \texttt{MAX}, \texttt{SUM},
  \texttt{AVG} \}$.} to column \texttt{agg\_col}.  \texttt{WHERE} can
have zero or multiple conditions, which means that column
\texttt{cond\_col} must satisfy the constraints expressed by the
operator \texttt{cond\_op}\footnote{\texttt{cond\_op} $\in
  \{=,<,>\}$.} and the condition value \texttt{cond}. Sketches for SQL
queries are simply the (sorted) sequences of
condition operators \texttt{cond\_op} in \texttt{WHERE} clauses.
For example, in Table~\ref{tbl:dataset}, sketch
``\texttt{WHERE}~\textgreater~\texttt{AND}~='' has two condition
operators, namely ``\textgreater'' and ``=''.

The generation of SQL queries differs from our previous semantic
parsing tasks, in that the table schema serves as input in addition to
natural language.  We therefore modify our input encoder in order to
render it table-aware, so to speak. Furthermore, due to the formulaic
nature of the SQL query, we only use our decoder to generate the
\texttt{WHERE} clause (with the help of sketches). The \texttt{SELECT}
clause has a fixed number of slots (i.e., aggregation operator
\texttt{agg\_op} and column \texttt{agg\_col}), which we
straightforwardly predict with softmax classifiers (conditioned on
the input). We briefly explain how these components are modeled below.

\begin{figure}[t]
\centering
\includegraphics[width=0.48\textwidth]{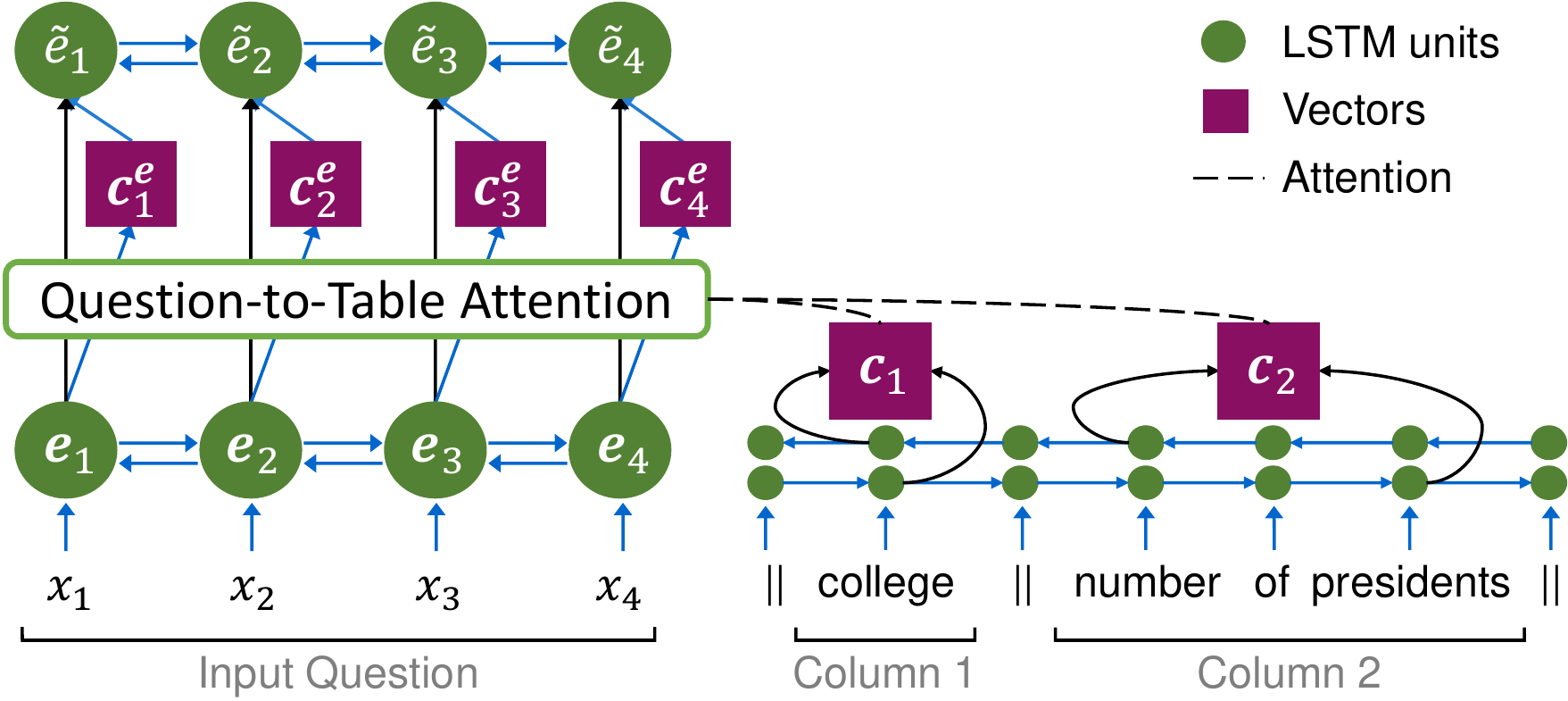}
\caption{Table-aware input encoder (left) and table column encoder (right) used for \textsc{WikiSQL}.}
\label{fig:wikisql:encoder}
\end{figure}

\paragraph{Table-Aware Input Encoder}
Given a table schema with $M$ columns, we employ the special
token ``\textbardbl'' to concatenate its header names as
``\textbardbl$c_{1,1} \cdots
c_{1,|c_1|}$\textbardbl$\cdots$\textbardbl$c_{M,1} \cdots
c_{M,|c_M|}$\textbardbl'', where the $k$-th column (``$c_{k,1} \cdots
c_{k,|c_k|}$'') has $|c_k|$ words. As shown in
Figure~\ref{fig:wikisql:encoder}, we use bi-directional LSTMs to
encode the whole sequence. Next, for column~$c_k$, the LSTM hidden
states at positions $c_{k,1}$ and $c_{k,|c_k|}$ are
concatenated. Finally, the concatenated vectors are used as the
encoding vectors $\{\mathbf{c}_k\}_{k=1}^{M}$ for table columns.

As mentioned earlier, the meaning representations of questions are
dependent on the tables.  As shown in
Figure~\ref{fig:wikisql:encoder}, we encode the input question~$x$
into $\{\mathbf{e}_t\}_{t=1}^{|x|}$ using LSTM units. At each time
step~$t$, we use an attention mechanism towards table column vectors
$\{\mathbf{c}_k\}_{k=1}^{M}$ to obtain the most relevant columns
for~$\mathbf{e}_t$. The attention score from $\mathbf{e}_t$ to
${\mathbf{c}}_{ k }$ is computed via ${ u }_{ t,k } \propto { \exp \{
  \alpha({\mathbf{e}}_{ t }) \cdot \alpha({\mathbf{c}}_{ k }) \} }$,
where $\alpha(\cdot)$ is a one-layer neural network, and $\sum _{ k=1
}^{ M }{ { u }_{ t,k } } = 1$. Then we compute the context vector
$\mathbf{c}_{t}^{e} = \sum_{ k=1 }^{ M }{ { u }_{ t,k } {\mathbf{c}}_{
    k } }$ to summarize the relevant columns for $\mathbf{e}_t$.  We
feed the concatenated vectors $\{[\mathbf{e}_{t} ,
\mathbf{c}_{t}^{e}]\}_{t=1}^{|x|}$ into a bi-directional LSTM encoder,
and use the new encoding vectors
$\{\tilde{\mathbf{e}}_t\}_{t=1}^{|x|}$ to replace
$\{\mathbf{e}_t\}_{t=1}^{|x|}$ in other model components.  We define
the vector representation of input~$x$ as:
\begin{equation}
\label{eq:wikisql:question}
\tilde{\mathbf{e}} = [\overrightarrow{\tilde{\mathbf{e}}}_{ |x| }, \overleftarrow{\tilde{\mathbf{e}}}_{ 1 }]
\end{equation}
analogously to \Crefrange{eq:encoder:lstm:right}{eq:encoder:lstm}.

\paragraph{\texttt{SELECT} Clause}
We feed the question vector $\tilde{\mathbf{e}}$ into a softmax
classifier to obtain the aggregation operator \texttt{agg\_op}.  If
\texttt{agg\_col} is the \mbox{$k$-th} table column, its probability
is computed via:
\begin{align}
\sigma(\mathbf{x}) = {\mathbf{w}_3 \cdot \tanh \left( \mathbf{W}_4 \mathbf{x} + \mathbf{b}_4 \right) } \label{eq:wikisql:agg_col:score} \\
p\left( \texttt{agg\_col}=k | x \right) \propto  { \exp \{ \sigma([ \tilde{\mathbf{e}} , {\mathbf{c}}_{ k } ]) \} } \label{eq:wikisql:agg_col}
\end{align}
where $\sum _{ j=1 }^{ M }{ p\left( \texttt{agg\_col} = j | x \right) } = 1$, $\sigma(\cdot)$ is a scoring network, and $\mathbf{W}_4 \in \mathbb{R}^{2n \times m} , \mathbf{w}_3 , \mathbf{b}_4 \in \mathbb{R}^{m}$ are parameters.

\paragraph{\texttt{WHERE} Clause}
We first generate sketches whose details are subsequently decorated by
the fine meaning decoder described in Section~\ref{sec:mr:generation}.
As the number of sketches in the training set is small ($35$~in
total), we model sketch generation as a classification problem.  We
treat each sketch~$a$ as a category, and use a softmax classifier to
compute~$p\left( a | x \right)$:
\begin{equation}
p\left( a | x \right) = \softmax_{a} \left( \mathbf{W}_a \tilde{\mathbf{e}} + \mathbf{b}_a \right) \nonumber
\end{equation}
where $\mathbf{W}_a \in \mathbb{R}^{|\mathcal{V}_a| \times n} ,
\mathbf{b}_a \in \mathbb{R}^{|\mathcal{V}_a|}$ are parameters, and
$\tilde{\mathbf{e}}$ is the table-aware input representation
defined in Equation~\eqref{eq:wikisql:question}.

\begin{figure}[t]
\centering
\includegraphics[width=0.46\textwidth]{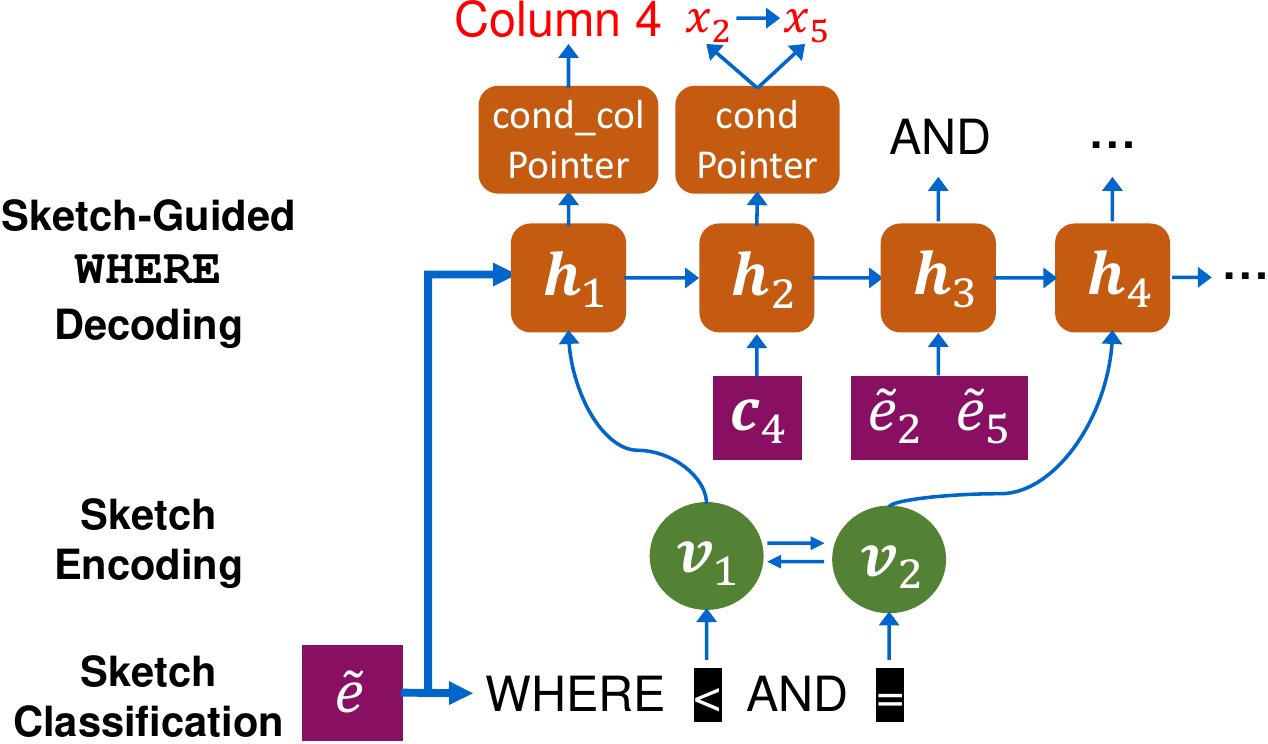}
\caption{Fine meaning decoder of the \texttt{WHERE} clause used for \textsc{WikiSQL}.}
\label{fig:wikisql:decoder}
\end{figure}

Once the sketch is predicted, we know
the condition operators and number of conditions in the \texttt{WHERE}
clause which follows the format ``\texttt{WHERE} (\texttt{cond\_op}
\texttt{cond\_col} \texttt{cond}) \texttt{AND} ...''. As shown in Figure~\ref{fig:wikisql:decoder}, our generation
task now amounts to populating the sketch with condition columns
\texttt{cond\_col} and their values \texttt{cond}.

Let $\{ \mathbf{h}_{t} \}_{t=1}^{|y|}$ denote the LSTM hidden states of the
fine meaning decoder, and $\{ \mathbf{h}_{t}^{att} \}_{t=1}^{|y|}$ the
vectors obtained by the attention mechanism
as in Equation~\eqref{eq:attention:new:hidden}.  The condition column
$\texttt{cond\_col}_{y_t}$ is selected from the table's headers. For
the \mbox{$k$-th} column in the table, we compute $p\left(
  \texttt{cond\_col}_{y_t}=k | y_{<t} , x , a \right)$ as in
Equation~\eqref{eq:wikisql:agg_col}, but use different parameters and
compute the score via $\sigma([ \mathbf{h}_{t}^{att} , {\mathbf{c}}_{
  k } ])$.  If the \mbox{$k$-th} table column is selected, we use
$\mathbf{c}_k$ for the input of the next LSTM unit in the decoder.

Condition values are typically mentioned in the input questions. These
values are often phrases with multiple tokens (e.g., \textit{Mikhail
  Snitko} in Table~\ref{tbl:dataset}). We therefore propose to
select a \emph{text span} from input~$x$ for each condition
value~$\texttt{cond}_{y_t}$ rather than copying tokens one by one.
Let $x_l \cdots x_r$ denote the text span from which
$\texttt{cond}_{y_t}$ is copied. We factorize its probability as:
\begin{align}
p&\left( \texttt{cond}_{y_t}=x_l \cdots x_r | y_{<t} , x , a \right) \nonumber \\
&=  p\left( \llbracket l \rrbracket_{y_t}^L | y_{<t} , x , a \right) p\left( \llbracket r \rrbracket_{y_t}^R | y_{<t} , x , a, \llbracket l \rrbracket_{y_t}^L \right) \nonumber \\
p&\left( \llbracket l \rrbracket_{y_t}^L | y_{<t} , x , a \right) \propto  { \exp \{ \sigma([ \mathbf{h}_{t}^{att} , \tilde{\mathbf{e}}_l ]) \} } \nonumber \\
p&\left( \llbracket r \rrbracket_{y_t}^R | y_{<t} , x , a, \llbracket l \rrbracket_{y_t}^L \right) \propto  { \exp \{ \sigma([ \mathbf{h}_{t}^{att} , \tilde{\mathbf{e}}_l , \tilde{\mathbf{e}}_r ]) \} } \nonumber
\end{align}
where $\llbracket l
\rrbracket_{y_t}^L / \llbracket r \rrbracket_{y_t}^R$ represents the
first/last copying index of $\texttt{cond}_{y_t}$ is $l/r$,
the probabilities are normalized to $1$, 
and $\sigma(\cdot)$ is the scoring network defined in
Equation~\eqref{eq:wikisql:agg_col:score}. Notice that we use
different parameters for the scoring networks $\sigma(\cdot)$.  The copied span is
represented by the concatenated vector $[\tilde{\mathbf{e}}_l ,
\tilde{\mathbf{e}}_r]$, which is fed into a one-layer neural network
and then used as the input to the next LSTM unit in the decoder.

\section{Experiments}

We present results on the three semantic parsing tasks discussed in
Section~\ref{sec:task}. Our implementation and pretrained models are available
at \url{https://github.com/donglixp/coarse2fine}.

\subsection{Experimental Setup}
\label{sec:experimental-setup}

\paragraph{Preprocessing}
For \textsc{Geo} and \textsc{Atis}, we used the preprocessed versions
provided by~\citet{lang2logic}, where natural language expressions are
lowercased and stemmed with NLTK~\cite{nltk}, and entity mentions are
replaced by numbered markers. We combined predicates and left brackets
that indicate hierarchical structures to make meaning representations
compact.  We employed the preprocessed \textsc{Django} data provided
by~\citet{nl2code}, where input expressions are tokenized by NLTK, and
quoted strings in the input are replaced with place holders.
\textsc{WikiSQL} was preprocessed by the script provided
by~\citet{wikisql}, where inputs were lowercased and tokenized by
Stanford CoreNLP~\cite{corenlp}.

\paragraph{Configuration}

Model hyperparameters were cross-validated on the training set for
\textsc{Geo}, and were validated on the development split for the
other datasets. Dimensions of hidden vectors and word embeddings were
selected from $\{ 250, 300 \}$ and $\{ 150, 200, 250, 300 \}$,
respectively. The dropout rate was selected from $\{ 0.3, 0.5 \}$.
Label smoothing~\cite{label:smoothing} was employed for \textsc{Geo}
and \textsc{Atis}. The smoothing parameter was set to $0.1$.
For \textsc{WikiSQL}, the hidden size of $\sigma(\cdot)$
and $\alpha(\cdot)$ in Equation~\eqref{eq:wikisql:agg_col:score}
was set to $64$.
Word embeddings were initialized by GloVe~\cite{glove}, and were
shared by table encoder and input encoder in
Section~\ref{sec:method:wikisql}.
We appended $10$-dimensional part-of-speech tag vectors to
embeddings of the question words in \textsc{WikiSQL}.
The part-of-speech tags were obtained by the spaCy toolkit.
We used the RMSProp optimizer~\cite{rmsprop} to train the models.
The learning rate was selected from $\{ 0.002, 0.005 \}$. The
batch size was $200$ for \textsc{WikiSQL}, and was $64$ for other
datasets. Early stopping was used to determine the number of epochs.

\paragraph{Evaluation} We use accuracy as the evaluation metric,
i.e.,~the percentage of the examples that are correctly parsed to
their gold standard meaning representations. For \textsc{WikiSQL}, we
also execute generated SQL queries on their corresponding tables, and
report the execution accuracy which is defined as the proportion of
correct answers.

\subsection{Results and Analysis}
\label{sec:results}

We compare our model (\ours) against several
previously published systems as well as various baselines.
Specifically, we report results with a model which decodes
meaning representations in one stage ({\textsc{OneStage}) without
leveraging sketches. We also report the results of several ablation
models, i.e., without a sketch encoder and
without a table-aware input encoder.

\begin{table}[t]
\centering
\small
\begin{tabular}{l c c}
	\toprule
	\textbf{Method}                     & \textbf{\textsc{Geo}} & \textbf{\textsc{Atis}} \\ \midrule
	\textsc{ZC07}~\cite{zc07}           & 86.1                  & 84.6                   \\
	\textsc{UBL}~\cite{ubl}             & 87.9                  & 71.4                   \\
	\textsc{FUBL}~\cite{fubl}           & 88.6                  & 82.8                   \\
	\textsc{GUSP++}~\cite{gusp}         & ---                     & 83.5                   \\
	\textsc{KCAZ13}~\cite{onthefly13}   & 89.0                  & ---                      \\
	\textsc{DCS+L}~\cite{dcs}           & 87.9                  & ---                      \\
	\textsc{TISP}~\cite{tisp}           & 88.9                  & 84.2                   \\ \midrule
	\textsc{Seq2Seq}~\cite{lang2logic}  & 84.6                  & 84.2                   \\
	\textsc{Seq2Tree}~\cite{lang2logic} & 87.1                  & 84.6                   \\
	\textsc{ASN}~\cite{asn}             & 85.7                  & 85.3                   \\
	\textsc{ASN+SupAtt}~\cite{asn}      & 87.1                  & 85.9                   \\ \midrule
	\base                               & 85.0                  & 85.3                   \\
	\ours                               & 88.2                  & 87.7                   \\
	~~~~$-$ sketch encoder              & 87.1                  & 86.9                   \\
	~~~~$+$ oracle sketch               & 93.9                  & 95.1                   \\ \bottomrule
\end{tabular}
\normalsize
\caption{Accuracies on \textsc{Geo} and \textsc{Atis}.}
\label{tbl:results:geo:atis}
\end{table}

Table~\ref{tbl:results:geo:atis} presents our results on
\textsc{Geo} and \textsc{Atis}.  Overall, we observe that
\ours~outperforms \base, which suggests that disentangling high-level
from low-level information during decoding is beneficial. The results
also show that removing the sketch encoder harms performance since the
decoder loses access to additional contextual information. Compared
with previous neural models that utilize syntax or grammatical
information (\textsc{Seq2Tree}, \textsc{ASN}; the second block in
Table~\ref{tbl:results:geo:atis}), our method performs competitively
despite the use of relatively simple decoders. As an upper bound, we
report model accuracy when gold meaning sketches are given to the fine
meaning decoder ($+$oracle sketch). As can be seen, predicting the
sketch correctly boosts performance.
The oracle results also indicate the accuracy of the fine meaning decoder.

\begin{table}[t]
	\centering
	\small
	\begin{tabular}{l c}
		\toprule
		\textbf{Method}                           & \textbf{Accuracy} \\ \midrule
		Retrieval System                          & 14.7              \\
		Phrasal SMT                               & 31.5              \\
		Hierarchical SMT                          & 9.5               \\ \midrule
		\textsc{Seq2Seq}+UNK replacement          & 45.1              \\
		\textsc{Seq2Tree}+UNK replacement         & 39.4              \\
		\textsc{LPN+Copy}~\cite{latent-predictor} & 62.3              \\
		\textsc{SNM+Copy}~\cite{nl2code}          & 71.6              \\ \midrule
		\base                                     & 69.5              \\
		\ours                                     & 74.1              \\
		~~~~$-$ sketch encoder                    & 72.1              \\
		~~~~$+$ oracle sketch                     & 83.0              \\ \bottomrule
	\end{tabular}
	\normalsize
	\caption{\textsc{Django} results. Accuracies in the first and second block are taken from~\citet{latent-predictor} and~\citet{nl2code}.}
	\label{tbl:results:django}
\end{table}

Table~\ref{tbl:results:django} reports results on \textsc{Django}
where we observe similar tendencies. \textsc{Coarse2Fine} outperforms
\textsc{OneStage} by a wide margin. It is also superior to the best
reported result in the literature (\textsc{snm}$+$\textsc{copy}; see the
second block in the table). Again we observe that the sketch encoder is
beneficial and that there is an $8.9$~point difference in accuracy between
\textsc{Coarse2Fine} and the oracle.

Results on \textsc{WikiSQL} are shown in
Table~\ref{tbl:results:wikisql}.  Our model is superior to
\textsc{OneStage} as well as to previous best performing systems.
\textsc{Coarse2Fine}'s accuracies on aggregation \texttt{agg\_op} and
\texttt{agg\_col} are $90.2\%$ and $92.0\%$, respectively, which is
comparable to \textsc{SQLNet}~\cite{sqlnet}. So the most gain is
obtained by the improved decoder of the \texttt{WHERE} clause.  We
also find that a table-aware input encoder is critical for doing well
on this task, since the same question might lead to different SQL
queries depending on the table schemas. Consider the question
``\textsl{how many presidents are graduated from A}''. The SQL query
over table ``\textit{\textbardbl President\textbardbl
  College\textbardbl}'' is ``\texttt{SELECT}
\texttt{COUNT}(\textit{President}) \texttt{WHERE} (\textit{College} =
\textit{A})'', but the query over table ``\textit{\textbardbl
  College\textbardbl Number of Presidents\textbardbl}'' would be
``\texttt{SELECT} \textit{Number of Presidents} \texttt{WHERE}
(\textit{College} = \textit{A})''.

\begin{table}[t]
\centering
\small
\begin{tabular}{l c c}
	\toprule
	\textbf{Method}                   & \textbf{Accuracy} & \textbf{\tabincell{c}{Execution \\
	Accuracy}} \\ \midrule
	\textsc{Seq2Seq}                  & 23.4              & 35.9                            \\
	Aug Ptr Network                   & 43.3              & 53.3                            \\
	\textsc{Seq2SQL}~\cite{wikisql}   & 48.3              & 59.4                            \\
	\textsc{SQLNet}~\cite{sqlnet}     & 61.3              & 68.0                            \\ \midrule
	\base                             & 68.8              & 75.9                            \\
	\ours                             & 71.7              & 78.5                            \\
	~~~~$-$ sketch encoder            & 70.8              & 77.7                            \\
	~~~~$-$ table-aware input encoder & 68.6              & 75.6                            \\
	~~~~$+$ oracle sketch             & 73.0              & 79.6                            \\ \bottomrule
\end{tabular}
\normalsize
\caption{Evaluation results on \textsc{WikiSQL}. Accuracies in the first block are taken from~\citet{wikisql} and~\citet{sqlnet}.}
\label{tbl:results:wikisql}
\end{table}

\begin{table}[t]
\centering
\small
\begin{tabular}{l c c c c}
	\toprule
	\textbf{Method} & \textbf{\textsc{Geo}} & \textbf{\textsc{Atis}} & \textbf{\textsc{Django}} & \textbf{\textsc{WikiSQL}} \\ \midrule
	\base           & 85.4                  & 85.9                   & 73.2                     & 95.4                      \\
	\ours           & 89.3                  & 88.0                   & 77.4                     & 95.9                      \\ \bottomrule
\end{tabular}
\normalsize
\caption{Sketch accuracy. For \base, sketches are extracted from the meaning representations it generates.}
\label{tbl:results:sketch}
\end{table}

We also examine the predicted sketches themselves in
Table~\ref{tbl:results:sketch}. We compare sketches generated by
\textsc{Coarse2Fine} against \textsc{OneStage}. The latter model
generates meaning representations without an intermediate sketch
generation stage. Nevertheless, we can extract sketches from the
output of \textsc{OneStage} following the procedures described in
Section~\ref{sec:task}. Sketches produced by \textsc{Coarse2Fine} are
more accurate across the board. This is not surprising because our
model is trained explicitly to generate compact meaning sketches.
Taken together
(Tables~\ref{tbl:results:geo:atis}--\ref{tbl:results:wikisql}), our
results show that better sketches bring accuracy gains on \textsc{Geo},
\textsc{Atis}, and \textsc{Django}.  On \textsc{WikiSQL}, the
sketches predicted by \textsc{Coarse2Fine} are marginally
better compared with \textsc{OneStage}.
Performance improvements on this task
are mainly due to the fine meaning decoder.  We conjecture that by
decomposing decoding into two stages, \ours~can better match table
columns and extract condition values without interference from the
prediction of condition operators. Moreover, the sketch provides a
canonical order of condition operators, which is beneficial for the
decoding process~\cite{order:matters16,sqlnet}.

\section{Conclusions}
\label{sec:conclusions}

In this paper we presented a coarse-to-fine decoding framework for neural
semantic parsing.  We first generate meaning sketches which abstract
away from low-level information such as arguments and variable names
and then predict missing details in order to obtain full meaning
representations. The proposed framework can be easily adapted to
different domains and meaning representations.  Experimental results
show that coarse-to-fine decoding improves
performance across tasks. In the future, we would like to apply the
framework in a weakly supervised setting, i.e.,~to learn semantic
parsers from question-answer pairs and to explore alternative ways of
defining meaning sketches.

\paragraph{Acknowledgments} We would like to thank Pengcheng Yin for
sharing with us the preprocessed version of the \textsc{Django}
dataset.  We gratefully acknowledge the financial support of the
European Research Council (award number 681760; Dong, Lapata) and the
AdeptMind Scholar Fellowship program (Dong).

\bibliography{sketch}
\bibliographystyle{acl_natbib}

\end{document}